\documentclass[10pt,twocolumn,letterpaper]{article}

\usepackage[pagenumbers]{cvpr}

\usepackage{graphicx}
\usepackage{amsmath}
\usepackage{amssymb}
\usepackage{booktabs}
\usepackage{multirow}
\usepackage{xcolor}
\usepackage{colortbl}
\usepackage{pifont}
\usepackage{caption}

\definecolor{Gray}{gray}{0.95}
\newcolumntype{a}{>{\columncolor{Gray}}c}

\newcommand{\cmark}{\ding{51}}

\renewcommand{\paragraph}[1]{\par\vspace{0.35em}\noindent{\bfseries #1}\hspace{0.45em}}

\usepackage[pagebackref,breaklinks,colorlinks]{hyperref}
\usepackage[capitalize]{cleveref}
\crefname{section}{Sec.}{Secs.}
\Crefname{section}{Section}{Sections}
\Crefname{table}{Table}{Tables}
\crefname{table}{Tab.}{Tabs.}

\newcommand*{\affaddr}[1]{#1}
\newcommand*{\affmark}[1][*]{\textsuperscript{#1}}

\usepackage[symbol]{footmisc}

\def\cvprPaperID{}
\def\confName{}
\def\confYear{}

\begin{document}

\title{SketchKeyAnime: Reference-anchored Sparse Key-Sketch Animation Synthesis}

\author{%
Meixi Li\affmark[1] \quad
Xianlin Zhang\affmark[1] \quad
Yue Zhang\affmark[1] \quad
Xueming Li\affmark[1]\\
\affaddr{\affmark[1]Beijing University of Posts and Telecommunications}
}

\twocolumn[{
\maketitle

\begin{center}
    \captionsetup{type=figure}
    \includegraphics[width=0.98\textwidth]{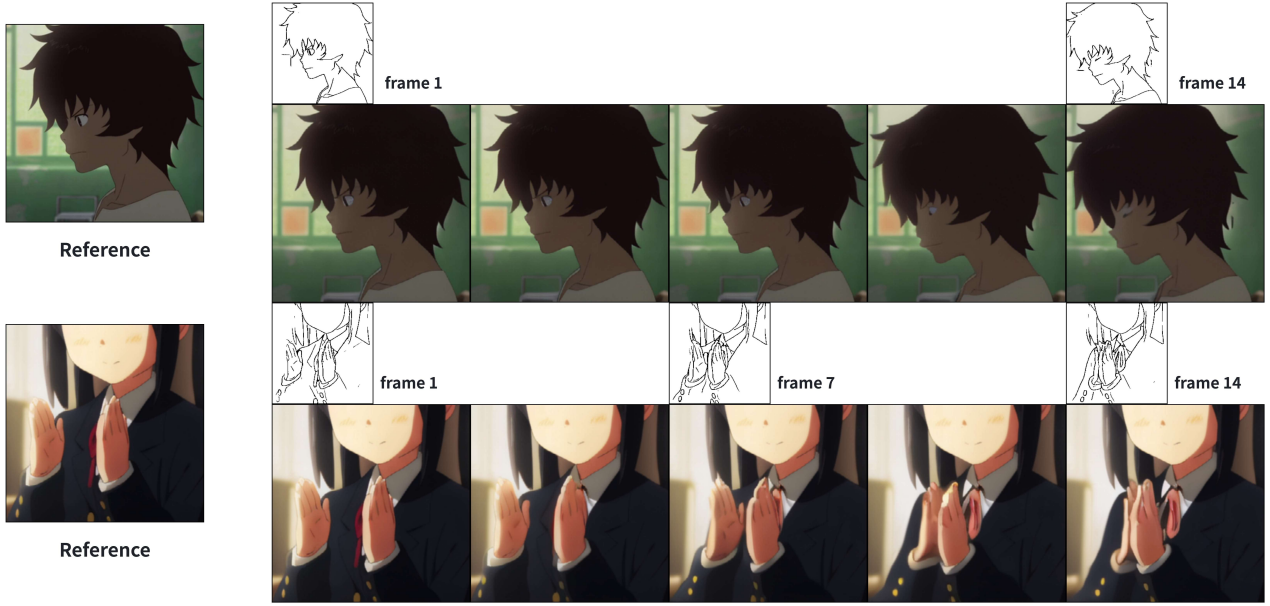}
    \captionof{figure}{
    Overview of the proposed reference-anchored sparse key-sketch animation synthesis task. Given a single RGB reference image and a small number of temporally indexed key sketches, SketchKeyAnime generates a complete animation sequence with consistent character appearance and controllable motion. Input images are from the ATD-12K dataset.
    }
    \label{fig:teaser}
\end{center}
}]

\begin{abstract}

Traditional animation production relies heavily on manual drawing and iterative refinement, particularly for key-pose design, in-betweening, and character coloring. While existing animation and video generation methods have made notable progress, they typically depend on RGB boundary frames, dense frame-wise conditions, or complete sketch sequences, limiting their applicability under low-cost input conditions. We present SketchKeyAnime, a video diffusion framework for generating structurally controllable, appearance-consistent, and temporally coherent animations from sparse key-sketch inputs. Given a single reference RGB image and a few temporally indexed key sketches, SketchKeyAnime introduces a dual-branch conditioning mechanism to encode local geometric constraints alongside semantic-temporal context. It leverages Sketch Cross Attention to fuse reference image and sketch conditions with learnable gating, and incorporates an Adaptive Weighted Loss to strengthen supervision on key-sketch frames and line-art regions. Experimental results on the Aesthetic subset of Sakuga-42M show that our approach consistently outperforms representative animation interpolation and sketch-guided generation baselines. Compared to the best-performing baseline, SketchKeyAnime reduces EDMD by 31.9\% and FVD by 9.5\%, demonstrating superior sketch fidelity and temporal coherence, while achieving the best overall performance across most quantitative metrics. These results validate the proposed framework and highlight its potential for low-cost, highly controllable animation creation.

\end{abstract}

\section{Introduction}

Animation is an important visual medium widely used in entertainment, education, and digital content creation. However, producing high-quality animation remains labor-intensive and time-consuming. A typical animation pipeline requires artists to design characters, draw key poses, complete in-between frames, and maintain consistent coloring and appearance across frames. In sketch-based workflows, artists often need to infer continuous motion from only a few key poses while preserving character structure and temporal consistency. Reducing such manual effort through automatic or semi-automatic animation generation is therefore of significant practical value.

Recent advances in generative models and diffusion models~\cite{ho2020ddpm,song2020ddim,rombach2022ldm,ho2022videodiffusion,blattmann2023svd,guo2024animatediff,wu2023tuneavideo,khachatryan2023text2videozero} have greatly promoted visual content generation and have inspired a variety of animation-related methods. 
Existing approaches can roughly be grouped into sketch-guided generation and video or animation inbetweening. 
Sketch-guided methods use sketches, line-art, or edges as structural conditions, but they often require dense frame-wise conditions, complete sketch sequences, text prompts, or specific input formats~\cite{zhang2023controlnet,guo2024sparsectrl,liu2025sketchvideo}. 
In contrast, interpolation-based methods synthesize intermediate frames from given boundary frames and are effective when RGB keyframes are available~\cite{li2023amt,xing2024tooncrafter,wang2025generativeinbetweening}. Nevertheless, animation data often contains large low-texture color regions, clear line boundaries, exaggerated pose changes, and nonlinear deformations, which can lead to broken lines, ghosting artifacts, or structural misalignment~\cite{li2021animeinterp,chen2022eisai,zhu2025tps}.
Overall, although existing methods have made progress in structural control and animation inbetweening, they remain limited in low-cost animation creation scenarios where users may only provide a reference appearance and a few key-pose sketches.

A practical sparse-sketch animation system should jointly satisfy three requirements. 
First, it should infer coherent intermediate motion from sparse structural constraints, since most frames do not have user-provided sketches. 
Second, it should preserve the character appearance from a reference RGB image while following the pose and structure specified by key sketches. 
Third, it should effectively learn from supervision concentrated on a small number of key-sketch frames and line-art regions. 
These requirements are difficult to satisfy simultaneously: frame-wise spatial control alone lacks temporal reasoning for frames without sketches, reference image and sketch conditions may compete during denoising, and uniform reconstruction losses may under-emphasize the sparse but important structural signals.

To address these issues, we propose SketchKeyAnime, a diffusion-based framework for reference-anchored sparse key-sketch animation synthesis. Given a single RGB reference image and a small set of temporally indexed key sketches, our method generates a complete animation sequence with consistent appearance, sketch-aligned structures, and coherent motion. 
We introduce a Dual-Branch Conditioning Mechanism that combines spatial key-sketch control with semantic-temporal sketch modeling, enabling both local structural guidance and motion reasoning across sparse sketches. We further design Sketch Cross Attention with learnable gating to balance reference-image appearance conditions and sketch-based structural guidance during denoising. Finally, we propose Adaptive Weighted Loss to emphasize key-sketch frames and line-art regions. Extensive experiments show that SketchKeyAnime outperforms representative animation interpolation and sketch-guided generation baselines in temporal consistency, appearance preservation, and motion fidelity.
Our main contributions are summarized as follows:
\begin{itemize}
    \item We introduce reference-anchored sparse key-sketch animation synthesis as a low-cost animation generation setting, and propose SketchKeyAnime to generate complete animation sequences from a single RGB reference image and a few temporally indexed key sketches.
    \item We design a Dual-Branch Conditioning Mechanism that exploits sparse key sketches through both frame-wise spatial control and semantic-temporal modeling, enhancing structural constraints at key-sketch frames and motion completion in frames without sketches.
    \item We propose Sketch Cross Attention, which fuses reference image conditions and sketch semantic conditions with learnable gating during diffusion denoising, balancing character appearance preservation and sketch-based structural control.
    \item We propose Adaptive Weighted Loss to strengthen key-sketch supervision at both frame and spatial levels, improving structural alignment and generation stability under sparse conditions.
\end{itemize}
\section{Related Work}
\label{sec:related_work}

\subsection{Sketch-guided Video Generation}

Sketches provide low-cost structural cues such as contours, poses, and spatial layouts, and have therefore been widely used for controllable visual generation. ControlNet~\cite{zhang2023controlnet} injects conditions such as sketches, edges, depth maps, and poses into pretrained diffusion models through an additional control branch, laying an important foundation for structure-controllable generation. With the development of video diffusion models, subsequent studies further extend such structural control to video generation, aiming to constrain both spatial content and cross-frame consistency~\cite{wang2023videocomposer}.
For controllable video diffusion, SparseCtrl~\cite{guo2024sparsectrl} introduces sparse condition control for text-to-video generation, enabling complete video generation from only a few sketch, depth, or RGB conditions, and modifies the vanilla ControlNet encoder for sparse-control scenarios. Motion-controllable generation methods further explore explicit motion guidance for video synthesis~\cite{wang2024motionctrl}. Ctrl-Adapter~\cite{lin2025ctrladapter} studies how to efficiently adapt existing ControlNets to different image and video diffusion backbones, supporting sparse-frame control and multiple structural conditions. These methods reduce the cost of dense frame-wise conditioning and validate the effectiveness of sparse structural signals in video generation. However, they mainly focus on incorporating structural conditions, while the joint modeling of reference appearance preservation, key-pose completion, and motion coherence in animation generation remains underexplored.

Research closer to sketch-based creation further explores generating dynamic content from hand-drawn sketches. 
Sketch Me A Video~\cite{zhang2021sketchmeavideo} alleviates the domain gap between free-hand sketches and real videos through retrieval and feature projection. 
SketchBetween~\cite{loftsdottir2022sketchbetween} targets sprite animation and generates in-between animations from rendered start/end keyframes and intermediate sketches. 
Recently, FlipSketch~\cite{bandyopadhyay2025flipsketch} generates hand-drawn-style animations from a single static sketch and a text description. 
VidSketch~\cite{jiang2026vidsketch} supports video animation generation from an arbitrary number of hand-drawn sketches and text prompts. 
SketchVideo~\cite{liu2025sketchvideo} enhances sketch-conditioned video generation and editing through sketch control blocks, inter-frame attention, and latent fusion. 
These works demonstrate the potential of sketches in video generation and animation creation, showing that hand-drawn structural conditions can provide intuitive and effective control.

In the animation domain, reference-based line-art video colorization methods~\cite{shi2023lineartcolorization,huang2024lvcd,meng2025anidoc} study how to generate temporally consistent colored animations under the guidance of reference images or character designs. Image animation methods such as MagicAnimate~\cite{xu2024magicanimate} and Animate Anyone~\cite{hu2024animateanyone} further study temporally consistent character animation from reference images. ToonCrafter~\cite{xing2024tooncrafter,xing2024dynamicrafter} introduces generative video priors into cartoon interpolation and uses line-art-related conditions to improve animation completion quality. These methods are closer to practical animation workflows, but they usually assume relatively complete frame-wise line-art sequences or explicit RGB boundary frames, which imposes high requirements on input completeness and limits flexible control with only a few key poses.

\subsection{Video Interpolation}

Video interpolation aims to synthesize intermediate frames from given adjacent or boundary frames, and is a classical task in video processing, slow-motion generation, and animation inbetweening. Traditional deep interpolation methods usually establish frame correspondences through optical flow estimation, feature matching, kernel prediction, or warping~\cite{niklaus2017sepconv,jiang2018superslomo,bao2019dain,lee2020adacof,kong2022ifrnet}. In recent years, RIFE~\cite{huang2022rife} achieves efficient real-time interpolation by directly estimating intermediate optical flows, FILM~\cite{reda2022film} improves interpolation quality under large motion through multi-scale features and large-motion modeling, and AMT~\cite{li2023amt} uses all-pairs correlation and multi-field transforms to better handle large motion and occlusion. These methods have achieved significant progress in natural video interpolation, but still rely on explicit correspondences or estimable motion fields between input RGB frames.
With the development of diffusion models, some methods reformulate video interpolation from a generative perspective~\cite{voleti2022mcvd}. LDMVFI~\cite{danier2024ldmvfi} formulates video frame interpolation as a conditional generation problem and leverages latent diffusion to improve perceptual quality. VIDIM~\cite{jain2024vidim} adopts cascaded diffusion models to generate complete intermediate videos given start and end frames. In addition, SEINE~\cite{chen2024seine} and Generative Inbetweening~\cite{wang2025generativeinbetweening} explore video diffusion models for transition generation or keyframe interpolation. These generative methods alleviate the reliance on precise motion correspondence.

Animation interpolation is more challenging than natural video interpolation. Animation videos often contain large textureless color regions, sharp line boundaries, exaggerated pose changes, and nonlinear deformations, causing natural video interpolation methods to suffer from line breakage, ghosting, or structural misalignment. To address these challenges, AnimeInterp~\cite{li2021animeinterp} systematically studies animation video interpolation and uses segment-guided matching and recurrent flow refinement to handle low-texture regions and large motion. EISAI~\cite{chen2022eisai} improves 2D animation interpolation from the perspective of perceptual quality, reducing line destruction and ghosting artifacts through forward warping and line-distance-related constraints. Thin-Plate Spline-based Interpolation for Animation Line Inbetweening~\cite{zhu2025tps} targets animation line-art inbetweening and uses thin-plate spline transformations to model large deformations between key line drawings. ToonCrafter~\cite{xing2024tooncrafter} further introduces generative video priors into cartoon interpolation, showing stronger animation completion ability under complex nonlinear motion and occlusion.

Overall, video and animation interpolation methods have made important progress in intermediate-frame completion, especially generative methods that improve the ability to handle complex motion. However, interpolation tasks usually assume that explicit RGB boundary frames are already provided, and the model mainly generates transitions between these known appearances and structures.

\section{Method}
\label{sec:method}

\begin{figure*}[!t]
    \centering
    \includegraphics[width=\textwidth]{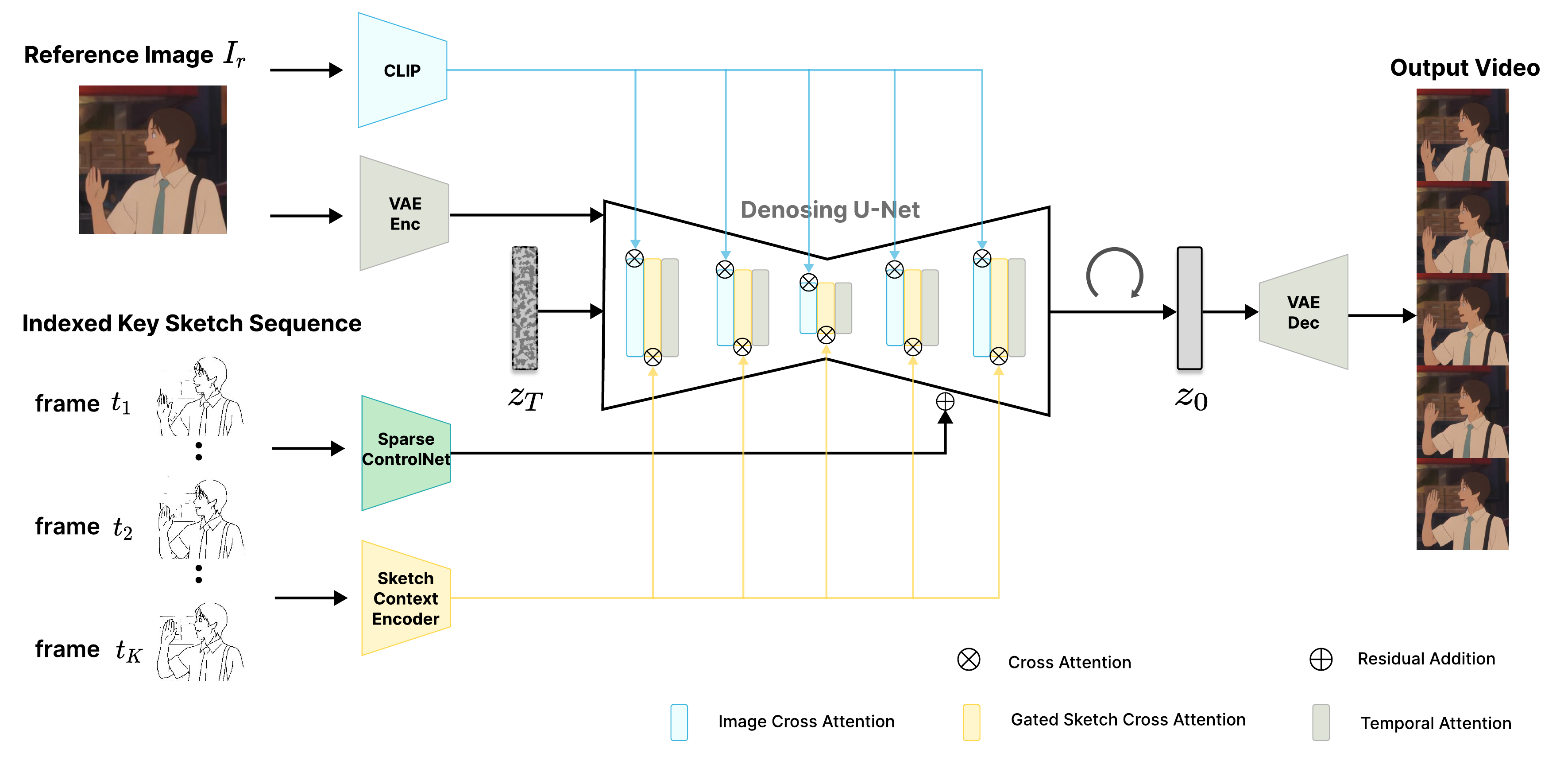}
    \caption{
    Overview of the SketchKeyAnime architecture. The framework takes a reference image and sparse temporally indexed key sketches as inputs. The reference image serves as the appearance condition, while the sparse key sketches provide structural and motion guidance through the Spatial Sparse ControlNet and the Semantic-Temporal Sketch Context Encoder. Conditioned on these complementary signals, the SVD-based denoising model generates a complete animation sequence with consistent character appearance and coherent motion. 
    }
    \label{fig:method_overview}
\end{figure*}

This section introduces the design of SketchKeyAnime. Given a single reference image \(I_r\) and a temporally indexed sparse key-sketch set \(\mathcal{S}=\{(S_k,t_k)\}_{k=1}^{K}\), where \(S_k\) denotes the \(k\)-th key sketch and \(t_k \in \{0,\dots,F-1\}\) denotes its corresponding video time position, the goal is to generate a complete animation sequence \(\mathcal{V}=\{I_t\}_{t=0}^{F-1}\).
As shown in Fig.~\ref{fig:method_overview}, SketchKeyAnime builds upon Stable Video Diffusion (SVD) as the base backbone and consists of three main modules: (1) the Dual-Branch Conditioning Mechanism exploits sparse key sketches from both spatial structural and semantic-temporal perspectives; (2) Sketch Cross Attention fuses the reference image condition and sketch semantic condition with learnable gating; and (3) Adaptive Weighted Loss strengthens the supervision on key-sketch frames and line-art regions.

\subsection{Preliminary}

Stable Video Diffusion (SVD)~\cite{blattmann2023svd} is a video generation framework based on latent diffusion models (LDMs)~\cite{rombach2022ldm,ho2020ddpm,song2020ddim,ho2022videodiffusion}. It performs diffusion modeling in the low-dimensional latent space constructed by a pretrained VAE~\cite{kingma2014vae} and uses a spatio-temporal UNet to model both per-frame visual content and cross-frame motion. Given an input video sequence \(x \in \mathbb{R}^{T \times C \times H \times W}\), SVD maps it into the latent space with a pretrained VAE encoder \(\mathcal{E}\), obtaining \(z_0 = \mathcal{E}(x) \in \mathbb{R}^{T \times c \times h \times w}\), where \(T\) denotes the number of video frames, \(C\) and \(c\) denote the channel dimensions in pixel and latent spaces, and \((H,W)\) and \((h,w)\) denote the original and latent spatial resolutions. In image-to-video generation, SVD incorporates reference conditions through both CLIP image features~\cite{radford2021clip} and the VAE latent of the reference image. The former is injected into the denoising network via cross-attention to provide semantic and appearance priors, while the latter is concatenated with the noisy video latent along the channel dimension to provide low-level spatial guidance. Given a clean video latent \(z_0\), the forward diffusion process produces a noisy latent \(z_\tau = \alpha_\tau z_0 + \sigma_\tau \epsilon,\ \epsilon \sim \mathcal{N}(0,I)\). In the reverse process, the denoising network predicts the clean latent representation from \(z_\tau\), diffusion timestep \(\tau\), and conditioning information \(c\), i.e., \(\hat{z}_0 = \epsilon_\theta(z_\tau,\tau,c)\). We express the SVD training objective using an equivalent latent-space denoising reconstruction form:
\begin{equation}
\label{eq:svd_loss}
\mathcal{L}_{\mathrm{SVD}}
=
\mathbb{E}_{z_0,\epsilon,\sigma,c}
\left[
w(\sigma)
\left\|
\hat{z}_0-z_0
\right\|_2^2
\right],
\end{equation}
where \(w(\sigma)\) is a reconstruction weight related to the noise strength.

\subsection{Dual-Branch Conditioning Mechanism}

Under sparse key-sketch conditioning, frame-wise spatial control can constrain keyframe structures but struggles to model motion relationships between key poses, while global semantic modeling helps capture motion context but is insufficient for accurate contour alignment at keyframes. To this end, we propose a \textbf{Dual-Branch Conditioning Mechanism}, which exploits key sketches from two complementary perspectives: local geometric constraints and global motion context.

\subsubsection{Spatial ControlNet Branch}

\begin{figure*}[!t]
    \centering
    \includegraphics[width=\linewidth]{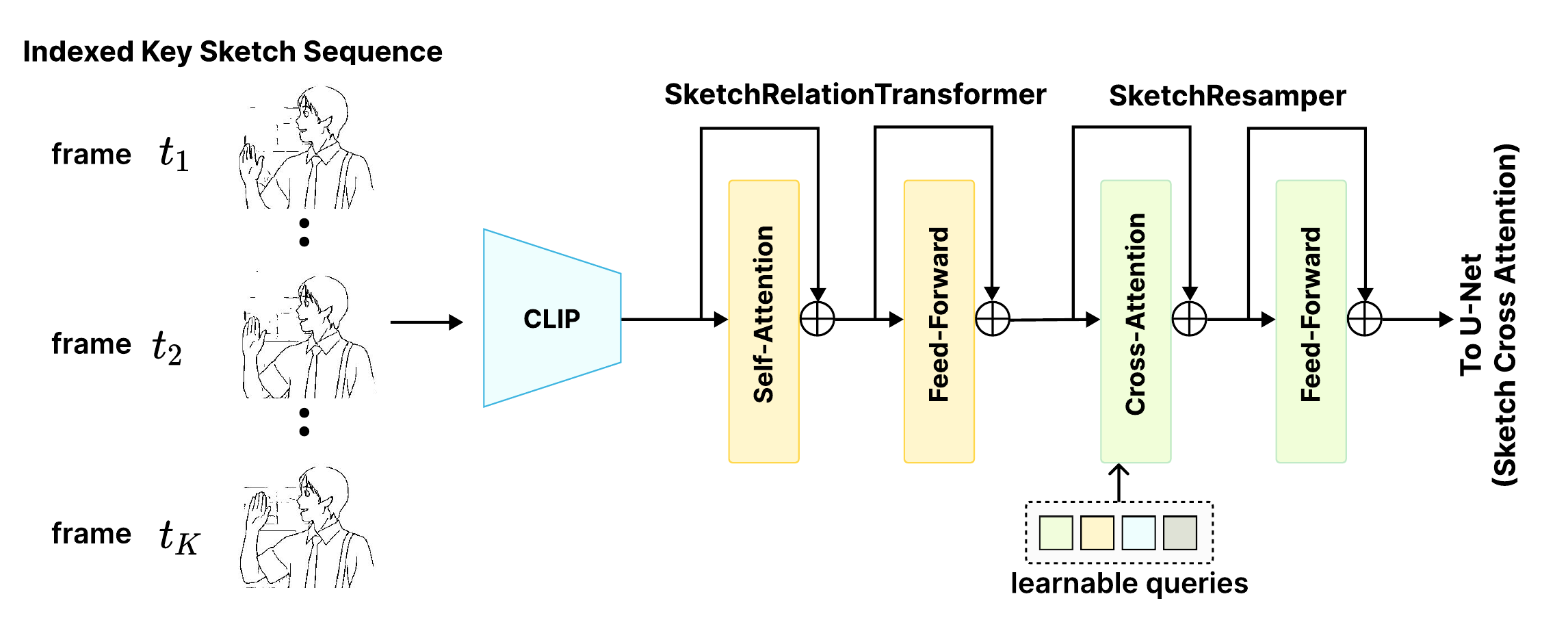}
    \caption{
    Semantic-Temporal Sketch Context Encoder. Given sparse key sketches with their temporal indices, a frozen CLIP image encoder first extracts semantic features from each sketch. The features are augmented with temporal positional embeddings and processed by the Sketch Relation Transformer to model semantic relationships and temporal dependencies among key sketches. The Sketch Resampler then converts the variable-length sketch features into fixed-format frame-wise sketch context tokens, which are injected into the Sketch Cross Attention modules of the SVD UNet.
    }
    \label{fig:sketchcontextencoder}
\end{figure*}

To provide direct local geometric constraints at key-sketch frames, we adopt a spatial ControlNet branch to inject sparse key sketches in a frame-wise manner. Let the temporal index set of key sketches be $\mathcal{T}=\{t_1,t_2,\dots,t_K\}$. We expand the sparse key-sketch set $\mathcal{S}$ into a sparse sketch sequence $\tilde{\mathcal{S}}=\{\tilde{S}_t\}_{t=0}^{F-1}$ of length $F$:
\begin{equation}
\label{eq:sparse_sketch_sequence}
\tilde{S}_t=
\begin{cases}
S_k, & t=t_k,\quad t\in\mathcal{T},\\
\mathbf{0}, & t\notin\mathcal{T}.
\end{cases}
\end{equation}
We also define a frame-wise conditioning mask:
\begin{equation}
\label{eq:frame_conditioning_mask}
M_t=
\begin{cases}
1, & t\in\mathcal{T},\\
0, & t\notin\mathcal{T}.
\end{cases}
\end{equation}
The control input for the spatial branch at frame $t$ is then
\begin{equation}
\label{eq:spatial_control_input}
c_t=[\tilde{S}_t,M_t],
\end{equation}
where $[\cdot,\cdot]$ denotes channel-wise concatenation. This unified input allows the model to distinguish real key-sketch frames from frames without sketches. The spatial branch encodes $\{c_t\}_{t=0}^{F-1}$ and outputs multi-scale control residuals to the video UNet, providing contour, pose, and local structural constraints at key-sketch frames.

The spatial control branch follows the sparse-control modification of SparseCtrl rather than the vanilla ControlNet design. Vanilla ControlNet feeds the noisy latent $z_{\tau,t}$ together with the condition signal into the control encoder, which may cause the branch to over-rely on the noisy sample at frames without sketches and weaken its response to actual sketch conditions. We therefore remove the direct input path from the noisy latent to the control encoder:
\begin{equation}
\label{eq:vanilla_control_input}
x_t^{\text{vanilla}}=E_{\text{cond}}(c_t)+z_{\tau,t},
\end{equation}
\begin{equation}
\label{eq:ours_control_input}
x_t^{\text{ours}}=E_{\text{cond}}(c_t).
\end{equation}
Here, $E_{\text{cond}}(\cdot)$ denotes the initial mapping of the condition signal. With this design, the spatial branch focuses on extracting structural control information from sparse sketches and the conditioning mask. Since this branch remains frame-wise by nature and cannot model motion relationships among key sketches, we further introduce a semantic-temporal branch as a complement.

\subsubsection{Semantic-Temporal Sketch Branch}

To model the motion correlations and temporal dependencies among sparse key sketches, we design a Semantic-Temporal Sketch Context Encoder, as shown in Fig.~\ref{fig:sketchcontextencoder}. It encodes a variable number of key sketches into fixed-format frame-wise sketch semantic tokens, which serve as the conditional context for the subsequent Sketch Cross Attention. The encoder consists of three main components: a frozen CLIP image encoder, a Sketch Relation Transformer, and a Sketch Resampler.

First, we use a frozen CLIP image encoder to extract a global semantic representation from each key sketch:
\begin{equation}
\label{eq:clip_sketch_feature}
f_k=E_{\mathrm{CLIP}}(S_k),\quad k=1,\dots,K.
\end{equation}
Here, $E_{\mathrm{CLIP}}$ follows the image encoding architecture of SVD and remains frozen during training. We then add temporal positional encoding according to the actual frame index of each sketch:
\begin{equation}
\label{eq:temporal_sketch_feature}
\bar{f}_k=f_k+\mathrm{PE}(t_k).
\end{equation}

The sketch features with temporal information are fed into the Sketch Relation Transformer based on self-attention~\cite{vaswani2017attention}. Based on self-attention along the key-sketch dimension, this module allows different key sketches to exchange information and captures semantic correlations and temporal dependencies among key poses:
\begin{equation}
\label{eq:sketch_relation_transformer}
\hat{\mathbf{F}}
=
E_{\mathrm{relation}}(\bar{f}_1,\bar{f}_2,\dots,\bar{f}_K).
\end{equation}
Here, $E_{\mathrm{relation}}$ denotes the Sketch Relation Transformer, and $\hat{\mathbf{F}}$ denotes the sketch features enriched with semantic relationships and temporal positional information.

Since the number of key sketches varies across samples, we further use a Sketch Resampler to convert $\hat{\mathbf{F}}$ into fixed-format frame-wise conditional tokens. This module adopts a query-based resampling structure, where learnable query tokens aggregate information from relation-aware sketch features through cross-attention. Let the target video length be $F$, the number of sketch queries per frame be $L$, and the token dimension be $D$. The Resampler output is
\begin{equation}
\label{eq:sketch_resampler_output}
C^{\mathrm{sketch}}
=
\mathrm{Resampler}(\hat{\mathbf{F}})
\in
\mathbb{R}^{F\times L\times D}.
\end{equation}
The sketch semantic condition for frame $t$ is
\begin{equation}
\label{eq:framewise_sketch_condition}
C_t^{\mathrm{sketch}}
=
C^{\mathrm{sketch}}[t]
\in
\mathbb{R}^{L\times D}.
\end{equation}
Since different temporal positions have different query tokens, the Sketch Resampler can aggregate different contextual representations for different target frames from the same set of key-sketch features, thereby organizing sparse sketch information along the full temporal axis. The subsequent Sketch Cross Attention uses $C_t^{\mathrm{sketch}}$ as the sketch condition for frame $t$, injecting the structural semantics and motion context of key sketches into the diffusion denoising process.
\subsection{Sketch Cross Attention}

After obtaining the reference-image condition and the sketch semantic-temporal condition, the model needs to preserve character appearance consistency while responding to sketch-defined structural changes during diffusion denoising. To this end, we propose Sketch Cross Attention, which adds an additional sketch-conditioning branch on top of the original image cross-attention and controls its injection strength with learnable gating.

Let $H$ denote the current hidden state of the UNet, $C_{\mathrm{image}}$ denote the reference-image condition, and $C_{\mathrm{sketch}}$ denote the sketch condition. We share the query projected from the hidden state, $Q=W_QH$, and compute the attention outputs for the image and sketch conditions separately:
\begin{equation}
\label{eq:image_cross_attention}
O_{\mathrm{image}}
=
\mathrm{Attn}(Q,K_{\mathrm{image}},V_{\mathrm{image}}),
\end{equation}
\begin{equation}
\label{eq:sketch_cross_attention_output}
O_{\mathrm{sketch}}
=
\mathrm{Attn}(Q,K_{\mathrm{sketch}},V_{\mathrm{sketch}}),
\end{equation}
where the image condition uses the original key and value projections of cross-attention, while the sketch condition uses additional independent key and value projections. The two condition outputs are fused before the output projection:
\begin{equation}
\label{eq:cross_attention_fusion}
O
=
O_{\mathrm{image}}
+
\lambda_{\mathrm{eff}}O_{\mathrm{sketch}},
\end{equation}
\begin{equation}
\label{eq:sketch_cross_attention}
\mathrm{CA}(H)
=
W_OO.
\end{equation}
Here, $\lambda_{\mathrm{eff}}$ controls the influence of the sketch condition in the current layer. Since different network layers have different demands for appearance preservation and structural control, a fixed sketch weight is insufficient for all levels. Therefore, we assign an independent learnable gating parameter $\alpha$ to each injected layer and define
\begin{equation}
\label{eq:learnable_gate}
\lambda_{\mathrm{eff}}
=
\lambda_{\mathrm{base}}\cdot(\tanh(\alpha)+1),
\end{equation}
where $\lambda_{\mathrm{base}}$ is the base scaling factor. When $\alpha=0$, the sketch condition has a stable non-zero contribution at the beginning of training; during training, the layer-wise $\alpha$ adaptively adjusts the sketch response strength, enabling a flexible balance between reference appearance preservation and sketch-based structural control.

In implementation, Sketch Cross Attention is inserted into all BasicTransformerBlocks with cross-attention in the SVD spatio-temporal UNet, allowing sketch conditions to participate in denoising at multiple spatial resolutions and semantic levels.

\subsection{Adaptive Weighted Loss}

Since sparse key sketches provide explicit structural supervision only at a few temporal positions, a standard uniform reconstruction loss may weaken the learning of key poses and line-art structures. To address this issue, we propose an \textbf{Adaptive Weighted Loss}, which increases the training weights of key supervised regions at both the frame and spatial levels.

\paragraph{Frame-level Adaptive Weighting.}
At the frame level, we increase the weights of conditioned frames according to the sparsity of key sketches in the current sample. Given a target video of $F$ frames and the frame-wise conditioning mask $M_t$, the number of conditioned frames is $K=\sum_{t=0}^{F-1}M_t$. The sample sparsity ratio is defined as
\begin{equation}
\label{eq:sparsity_ratio}
r_{\mathrm{sparse}}
=
1-\frac{K}{F}.
\end{equation}
A smaller number of key sketches leads to a larger $r_{\mathrm{sparse}}$, and thus conditioned frames should receive stronger supervision. We define the frame-level adaptive coefficient as
\begin{equation}
\label{eq:frame_adaptive_coefficient}
\lambda_{\mathrm{frame}}
=
1+\lambda_1 r_{\mathrm{sparse}},
\end{equation}
where $\lambda_1$ is a hyperparameter. The corresponding frame-level weight is
\begin{equation}
\label{eq:frame_weight}
W_{\mathrm{frame}}(t)
=
1+(\lambda_{\mathrm{frame}}-1)M_t
=
\begin{cases}
\lambda_{\mathrm{frame}}, & M_t=1,\\
1, & M_t=0.
\end{cases}
\end{equation}
That is, frames with sketch conditions are assigned higher weights, while frames without sketch conditions retain the original weight.

\paragraph{Sketch-aware Spatial Weighting.}
At the spatial level, we further emphasize sketch line regions. Let $M_{\mathrm{line}}(t,x,y)$ denote the line-region mask obtained from the input sketch condition $\tilde{S}_t$, where a value of 1 indicates a line region and 0 indicates a background region. The spatial weight is defined as
\begin{equation}
\label{eq:line_weight}
W_{\mathrm{line}}(t,x,y)
=
\begin{cases}
\lambda_2, & M_t=1,\ M_{\mathrm{line}}(t,x,y)=1,\\
1, & M_t=1,\ M_{\mathrm{line}}(t,x,y)=0,\\
1, & M_t=0,
\end{cases}
\end{equation}
where $\lambda_2$ controls the enhancement strength of line-art regions. This weight only takes effect on line regions in frames with sketch conditions; for frames without sketch conditions, the spatial weight remains 1.

The total weight is
\begin{equation}
\label{eq:total_weight}
W_{\mathrm{total}}(t,x,y)
=
W_{\mathrm{frame}}(t)
\cdot
W_{\mathrm{line}}(t,x,y).
\end{equation}
The Adaptive Weighted Loss is defined as
\begin{equation}
\label{eq:adaptive_weighted_loss}
\begin{aligned}
\mathcal{L}_{\mathrm{adaptive}}
&=
\mathbb{E}_{\sigma,t,x,y}\Big[
w(\sigma)
\left(\hat{z}_0(t,x,y)-z_0(t,x,y)\right)^2 \\
&\quad \cdot W_{\mathrm{total}}(t,x,y)
\Big].
\end{aligned}
\end{equation}
This loss jointly adjusts the supervision strength according to sample sparsity, key-sketch frames, and line-art regions, encouraging the model to focus more on key structures under sparse conditions.

\section{Experiments}
\label{sec:experiments}

\subsection{Implementation Details}

\paragraph{Dataset.}
We use the Aesthetic subset of the Sakuga-42M dataset as the source for training and evaluation~\cite{pan2024sakuga42m}. This subset contains approximately 60k animation clips. We randomly select over 100 samples as the test set and use the remaining samples for training. To avoid abrupt content changes caused by shot transitions, we use PySceneDetect~\cite{pyscenedetect} for scene cut detection and construct video clips only within the same continuous scene. Each sample contains 15 temporal positions, with all frames resized to $256 \times 256$. The first frame is used as the RGB reference image, and the following 14 frames are used as the target animation sequence. Sketch conditions are extracted by Informative Drawings~\cite{chan2022informative} and then binarized.

\paragraph{Training strategy.}
SketchKeyAnime is trained based on Stable Video Diffusion~\cite{blattmann2023svd} in three stages. In the first stage, we train the Spatial ControlNet Branch for 50k steps to obtain basic sketch-based geometric responsiveness, where the first 20k steps use frame-wise sketch conditions and the remaining 30k steps use sparse key-sketch conditions. In the second stage, we train the Sketch Context Encoder and Sketch Cross Attention for 50k steps, enabling the model to extract semantic-temporal context from sparse key sketches and inject it into the denoising process. In the third stage, we further fine-tune Sketch Cross Attention and jointly optimize part of the spatial-attention-related UNet parameters with a small learning rate for 20k steps, improving the coordination between sketch semantic conditions and the SVD backbone.

\paragraph{Key-sketch sampling strategy.}
During training, we adopt a mixed key-sketch sampling strategy to improve adaptability to different sparse input configurations. It contains four modes: Bisection Mode preserves the first and last key-sketch frames and recursively selects temporal midpoints; Boundary Mode preserves the first and last frames and randomly adds key-sketch frames in the middle; Local Dense Mode densely samples key-sketch frames within local rapid-motion segments; and Random Mode randomly selects the number and positions of key sketches. The sampling ratio of the four modes is set to 6:2:1:1 throughout training.

\paragraph{Training and hyperparameters.}
All experiments are conducted on 4 NVIDIA V100 GPUs with a total batch size of 16, using AdamW as the optimizer. The depths of the Sketch Relation Transformer and Sketch Resampler are both set to 4, and the Resampler uses 16 sketch tokens per frame. The base sketch scaling factor $\lambda_{\mathrm{base}}$ in Sketch Cross Attention is set to 1. For Adaptive Weighted Loss, the frame-level coefficient $\lambda_1$ and the line region enhancement coefficient $\lambda_2$ are set to 1.5 and 2, respectively.

\subsection{Quantitative Comparisons}

\begin{table}[!t]
\centering
\caption{
Quantitative comparison with representative baselines under sparse key-sketch conditions. Results are averaged across three sparse-input settings using 2, 3, and 5 key sketches. For baselines that require RGB inputs, the input key sketches are colorized by MangaNinja using the same reference image before being fed to the corresponding methods.
}
\label{tab:quantitative_comparisons}
\resizebox{\linewidth}{!}{
\begin{tabular}{lccccc}
\toprule
Metric & SparseCtrl & ToonCrafter & AMT & EISAI & Ours \\
\midrule
EDMD $\downarrow$ & 7.2321 & 6.1119 & 5.9788 & 5.9700 & \textbf{4.1588} \\
FID $\downarrow$ & 82.3274 & \textbf{47.3319} & 90.3480 & 90.1897 & 65.5113 \\
FVD $\downarrow$ & 1109.4439 & 422.7736 & 768.6147 & 607.9490 & \textbf{382.5594} \\
LPIPS $\downarrow$ & 0.3415 & 0.1885 & 0.3273 & 0.2932 & \textbf{0.1861} \\
PSNR $\uparrow$ & 13.8070 & 19.3532 & 17.0297 & 17.0560 & \textbf{20.6994} \\
SSIM $\uparrow$ & 0.5144 & 0.6987 & 0.6207 & 0.6215 & \textbf{0.7058} \\
\bottomrule
\end{tabular}
}
\end{table}

\paragraph{Baselines.}
We compare SketchKeyAnime with four representative methods, including the cartoon video interpolation method ToonCrafter~\cite{xing2024tooncrafter}, the animation image interpolation method EISAI~\cite{chen2022eisai}, the general video interpolation method AMT~\cite{li2023amt}, and the sparse conditional video generation method SparseCtrl~\cite{guo2024sparsectrl}. Since some baselines cannot directly take sketches as input, we use MangaNinja~\cite{liu2025manganinja} to colorize the key sketches according to the same RGB reference image, and use the resulting colored keyframes as inputs for the corresponding baselines. For SparseCtrl, we use the colorized keyframes as sparse RGB conditions and leave the text condition empty to avoid insufficient appearance constraints from text prompts.

\begin{table}[!t]
\centering
\caption{
User study results measured by subjective preference rates. For RGB-input baselines, key sketches are colorized by MangaNinja using the same reference image. Percentages indicate voting rates for each criterion.
}
\label{tab:user_study}
\resizebox{\linewidth}{!}{
\begin{tabular}{lccccc}
\toprule
Criterion & SparseCtrl & ToonCrafter & AMT & EISAI & Ours \\
\midrule
Overall Quality $\uparrow$ & 6.06\% & 23.23\% & 6.06\% & 13.13\% & \textbf{51.52\%}\\
Sketch Alignment $\uparrow$ & 9.09\% & 22.22\% & 4.04\% & 10.10\% & \textbf{54.55\%} \\
Motion Quality $\uparrow$ & 13.13\% & 20.20\% & 8.08\% & 5.05\% & \textbf{53.54\%} \\
Temporal Coherence $\uparrow$ & 5.05\% & 17.17\% & 7.07\% & 11.11\% & \textbf{59.60\%} \\
\bottomrule
\end{tabular}
}
\end{table}

\begin{table*}[!t]
\centering
\caption{
Ablation study of the main components in SketchKeyAnime. All variants are evaluated under the same sparse key-sketch setting.
}
\label{tab:ablation}
\resizebox{\textwidth}{!}{
\begin{tabular}{lcccc|cccccc}
\toprule
Variant & Spatial Branch & AWL & S-T Sketch Branch & Gate
& EDMD $\downarrow$ & FID $\downarrow$ & FVD $\downarrow$ & LPIPS $\downarrow$ & PSNR $\uparrow$ & SSIM $\uparrow$ \\
\midrule
I. Spatial-only & \cmark &  &  & - 
& \textbf{3.9819} & 66.2395 & 424.8304 & 0.1957 & 20.1119 & 0.6924 \\
II. Spatial + AWL & \cmark & \cmark &  & - 
& 4.1239 & 67.2228 & 420.8799 & 0.2009 & 20.2114 & 0.6944 \\
III. Spatial + S-T Sketch Branch & \cmark &  & \cmark & Learnable 
& 4.1233 & 66.2830 & 385.5657 & 0.1887 & 20.6019 & 0.7028 \\
IV. Fixed-gate SketchKeyAnime & \cmark & \cmark & \cmark & Fixed 
& 4.1130 & 66.8089 & 407.4537 & 0.1912 & 20.6305 & 0.7002 \\
V. SketchKeyAnime & \cmark & \cmark & \cmark & Learnable 
& 4.1588 & \textbf{65.5113} & \textbf{382.5594} & \textbf{0.1861} & \textbf{20.6994} & \textbf{0.7058} \\
\bottomrule
\end{tabular}
}
\end{table*}

\paragraph{Metrics.}
We conduct quantitative evaluation from three aspects: 1) image and video generation quality, where FID~\cite{heusel2017fid} and FVD~\cite{unterthiner2019fvd} measure frame-level image distribution and video temporal distribution; 2) frame-level similarity to ground-truth animations, where LPIPS~\cite{zhang2018lpips}, PSNR, and SSIM~\cite{wang2004ssim} measure perceptual difference, pixel accuracy, and structural similarity; and 3) sketch structural alignment, where we adopt EDMD following LVCD~\cite{huang2024lvcd}. EDMD compares the Euclidean Distance Map Difference between the line-art extracted from generated frames and the input sketches. A lower EDMD indicates better structural alignment with the input sketches.

\paragraph{Results.}
In quantitative experiments, we evaluate cases with 2, 3, and 5 key sketches under recursive bisection sampling, and report the average results over the three settings. Although the optimal key-sketch positions may vary across animation clips, this setting provides unified and reproducible sparse input conditions for fair comparison. As shown in Table~\ref{tab:quantitative_comparisons}, SketchKeyAnime achieves the best results on most metrics except FID. In particular, it significantly outperforms all baselines in FVD, indicating that our method generates videos that are closer to the real animation distribution and more temporally coherent under sparse key-sketch conditions. Although MangaNinja+ToonCrafter achieves a lower FID, this may be related to ToonCrafter's toon rectification learning and detail-recovery decoder designed for cartoon interpolation, which can improve frame-level visual distribution quality in the cartoon domain. In addition, FID mainly measures frame-level image distribution and cannot fully reflect whether the generated results follow the input key sketches or preserve coherent cross-frame motion.

\subsection{Qualitative Comparisons}

\begin{figure*}[t]
    \centering
    \includegraphics[width=0.96\linewidth]{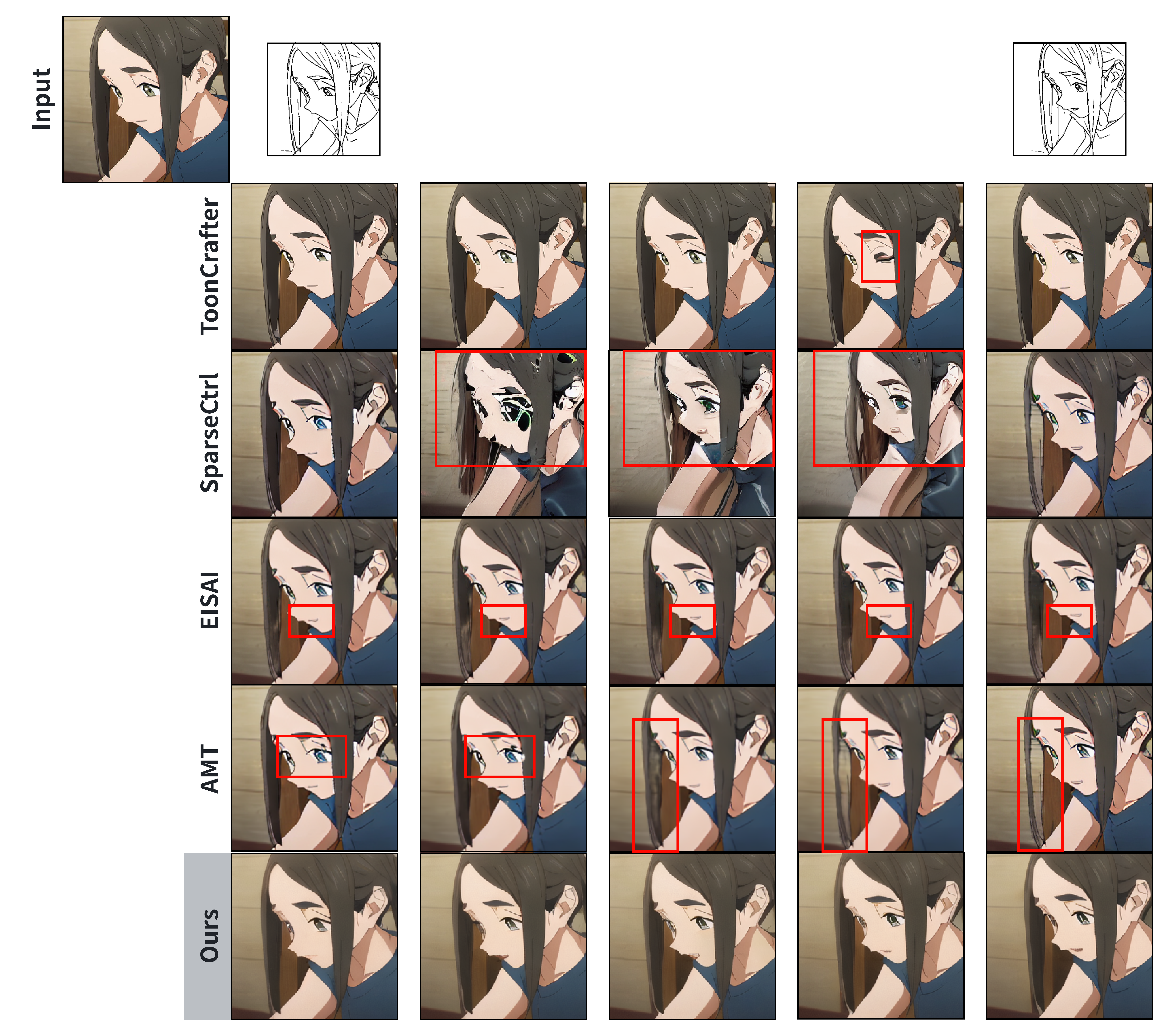}
    \caption{
    Qualitative comparison with SparseCtrl, ToonCrafter, AMT, and EISAI under sparse key-sketch conditions. For baselines that require RGB inputs, the input key sketches are first colorized by MangaNinja using the same reference image. This example uses key sketches from the first and last frames.
    }
    \label{fig:qualitative_comparison_first_last}
\end{figure*}

\begin{figure*}[t]
    \centering
    \includegraphics[width=0.96\linewidth]{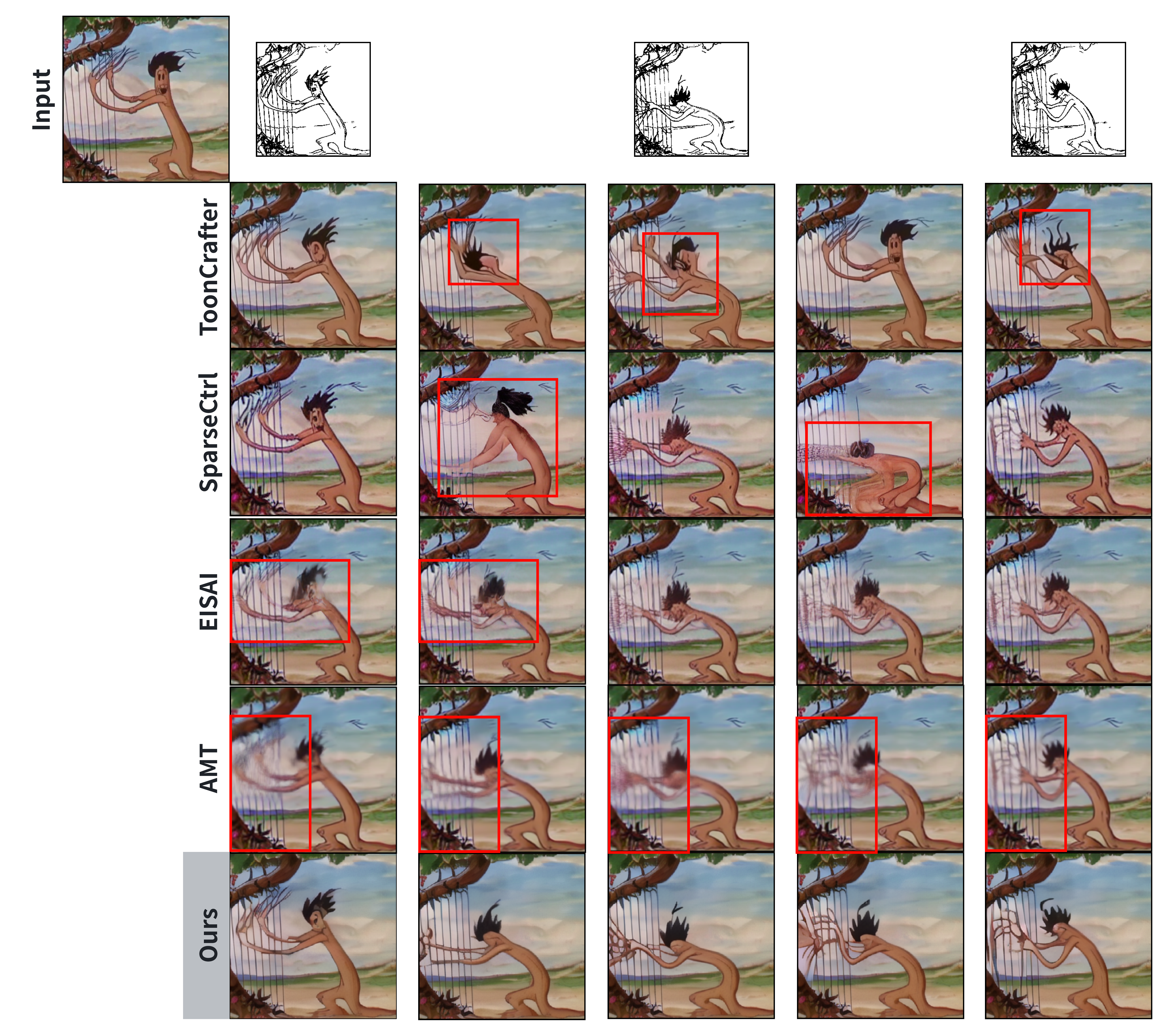}
    \caption{
    Qualitative comparison with SparseCtrl, ToonCrafter, AMT, and EISAI under sparse key-sketch conditions. For baselines that require RGB inputs, the input key sketches are first colorized by MangaNinja using the same reference image. This example uses key sketches from the first, middle, and last frames.
    }
    \label{fig:qualitative_comparison_first_middle_last}
\end{figure*}

The qualitative results are shown in Figs.~\ref{fig:qualitative_comparison_first_last} and~\ref{fig:qualitative_comparison_first_middle_last}. Compared with interpolation-based methods such as EISAI and AMT, our method does not simply produce uniform transitions between key sketches, but instead infers more reasonable intermediate motions from sparse key sketches, leading to more natural and smooth pose changes. Meanwhile, AMT tends to produce blurry results in some cases. Compared with SparseCtrl, SketchKeyAnime can more stably utilize the reference image and key-sketch conditions, avoiding severe line-structure changes, brightness flickering, and implausible motion variations. ToonCrafter can generate rich dynamic details, while our method shows more stable motion organization and line continuity under key-sketch constraints. Overall, SketchKeyAnime achieves a better balance among motion plausibility, transition smoothness, and line structure stability.

\subsection{User Study}

In addition to objective metrics, we conduct a user study to evaluate the subjective generation quality of different methods. We invite 11 participants and select 9 samples from the test set for evaluation. For each sample, we provide 2, 3, and 5 key sketches under recursive bisection sampling, and compare our method with the same baselines used in the quantitative experiments. All method names are hidden during evaluation. Participants are asked to rate the results from four aspects: overall quality, sketch alignment, motion quality, and temporal coherence. The user study results are shown in Table~\ref{tab:user_study}. Our method receives higher preference across all subjective criteria, further validating its effectiveness for sparse key-sketch animation synthesis.

\subsection{Ablation Study}

To analyze the effect of each module, we construct five variants under the same training data and evaluation setting:
\textbf{(I) Spatial-only}: only the Spatial ControlNet Branch is used.
\textbf{(II) Spatial + AWL}: Adaptive Weighted Loss is added to Spatial-only.
\textbf{(III) Spatial + S-T Sketch Branch}: the Semantic-Temporal Sketch Context Encoder and Sketch Cross Attention are added to Spatial-only.
\textbf{(IV) Fixed-gate SketchKeyAnime}: AWL and the semantic-temporal sketch branch are used, but the gate in Sketch Cross Attention is fixed.
\textbf{(V) SketchKeyAnime}: the full model with AWL, the semantic-temporal sketch branch, and learnable gating.

\begin{figure*}[t]
    \centering
    \includegraphics[width=0.96\linewidth]{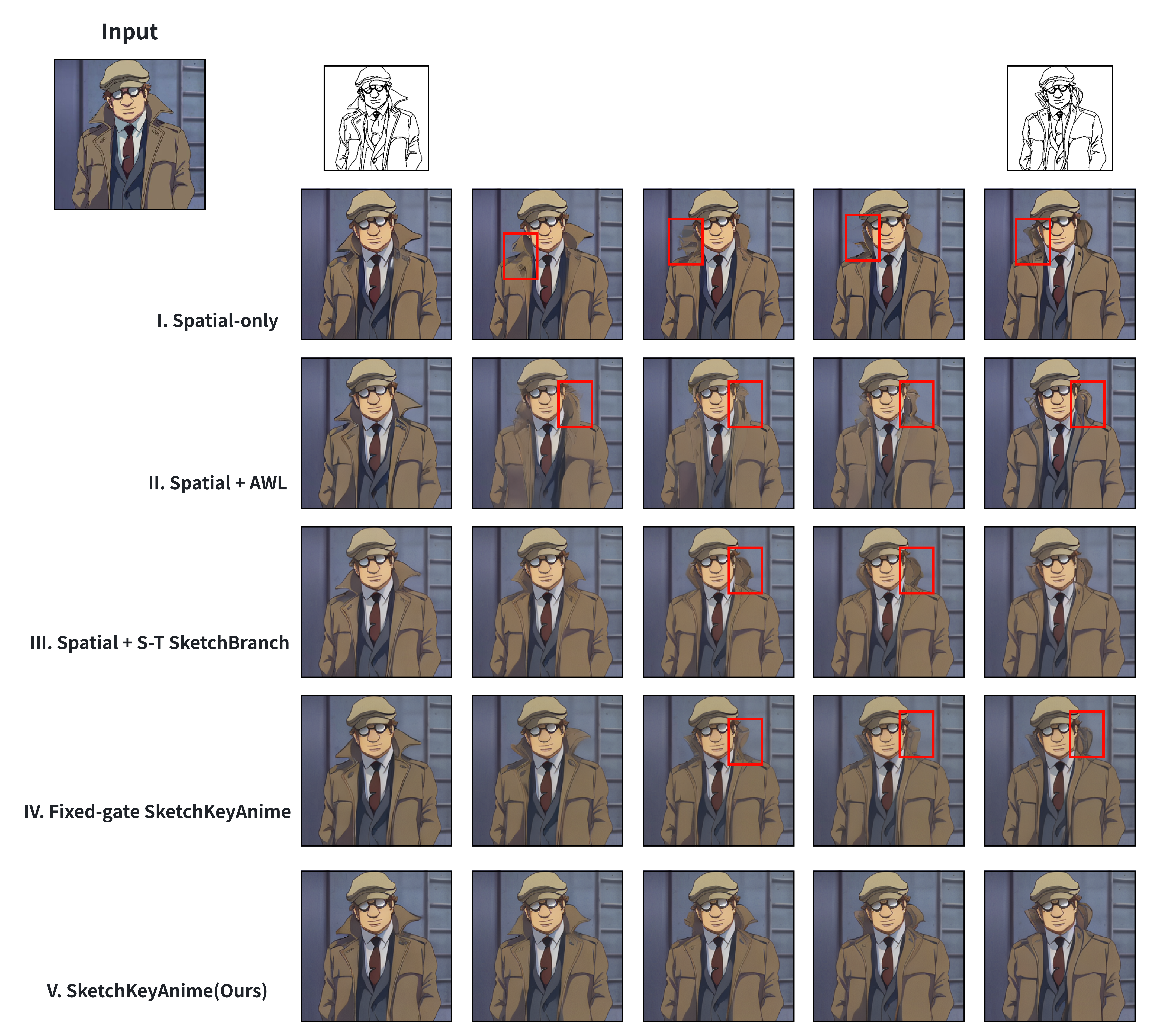}
    \caption{
    Qualitative ablation comparison of SketchKeyAnime components under sparse key-sketch conditions. The compared variants examine the effects of Adaptive Weighted Loss, the Semantic-Temporal Sketch Branch, and the learnable gating mechanism, with the Spatial ControlNet Branch used as the base setting.
    }
    \label{fig:ablation}
\end{figure*}

As shown in Table~\ref{tab:ablation}, Spatial-only achieves the lowest EDMD, indicating that the spatial branch already provides strong local structural constraints at key-sketch frames. However, its FVD is relatively high, suggesting that frame-wise spatial control alone is insufficient for coherent intermediate motion. In Fig.~\ref{fig:ablation}, this variant produces visible lines around the character's collar, but the motion changes are often abrupt.

Adding AWL slightly improves PSNR and SSIM, showing that weighting key supervised regions helps improve reconstruction quality and structural preservation. However, since AWL does not introduce temporal context, its benefit for motion completion is limited. In contrast, adding the Semantic-Temporal Sketch Branch reduces FVD from 424.8304 to 385.5657 and also improves LPIPS, PSNR, and SSIM, demonstrating that semantic-temporal sketch context provides useful motion cues for frames without sketch inputs and leads to more reasonable transitions.

Fixed-gate SketchKeyAnime uses both AWL and the semantic-temporal branch, but a fixed gate cannot adaptively balance the reference-image condition and the sketch condition, which may cause weak line control and blurred boundaries. The full model with learnable gating achieves the best FVD, LPIPS, PSNR, and SSIM, and produces clearer collar lines and more natural motion transitions in Fig.~\ref{fig:ablation}.

Overall, the ablation study shows that the modules are complementary: the spatial branch provides local geometric constraints, AWL strengthens key supervised regions, the Semantic-Temporal Sketch Branch improves motion completion, and learnable gating further balances reference appearance preservation and sketch-based structural control. The full model therefore achieves the best overall performance.

\section{Conclusion}

In this paper, we propose SketchKeyAnime, a framework for reference-anchored sparse key-sketch animation synthesis. To address the difficulty of balancing structural control, motion completion, and appearance preservation under sparse key-sketch conditions, we build upon Stable Video Diffusion and design a dual-branch conditioning mechanism. The Spatial ControlNet Branch provides keyframe-level geometric constraints, while the Semantic-Temporal Sketch Context Encoder models motion relationships among key sketches. Meanwhile, Sketch Cross Attention fuses reference image conditions and sketch conditions through learnable gating, and Adaptive Weighted Loss further strengthens supervision on key-sketch frames and line-art regions. Experimental results show that SketchKeyAnime can generate animation sequences with accurate structures, coherent motions, and consistent appearance from only a few key sketches, providing an effective solution for low-cost and highly controllable animation creation.

{\small
\bibliographystyle{ieee_fullname}
\bibliography{egbib}

\begin{thebibliography}{10}\itemsep=-1pt

\bibitem{bandyopadhyay2025flipsketch}
Hmrishav Bandyopadhyay and Yi-Zhe Song.
\newblock Flipsketch: Flipping static drawings to text-guided sketch animations.
\newblock In {\em Proceedings of the Computer Vision and Pattern Recognition Conference (CVPR)}, pages 28394--28404, June 2025.

\bibitem{bao2019dain}
Wenbo Bao, Wei-Sheng Lai, Chao Ma, Xiaoyun Zhang, Zhiyong Gao, and Ming-Hsuan Yang.
\newblock Depth-aware video frame interpolation.
\newblock In {\em Proceedings of the IEEE/CVF Conference on Computer Vision and Pattern Recognition (CVPR)}, pages 3703--3712, 2019.

\bibitem{blattmann2023svd}
Andreas Blattmann, Tim Dockhorn, Sumith Kulal, Daniel Mendelevitch, Maciej Kilian, Dominik Lorenz, Yam Levi, Zion English, Vikram Voleti, Adam Letts, Varun Jampani, and Robin Rombach.
\newblock Stable video diffusion: Scaling latent video diffusion models to large datasets.
\newblock {\em arXiv preprint arXiv:2311.15127}, 2023.

\bibitem{chan2022informative}
Caroline Chan, Fr\'edo Durand, and Phillip Isola.
\newblock Learning to generate line drawings that convey geometry and semantics.
\newblock In {\em Proceedings of the IEEE/CVF Conference on Computer Vision and Pattern Recognition (CVPR)}, pages 7915--7925, June 2022.

\bibitem{chen2022eisai}
Shuhong Chen and Matthias Zwicker.
\newblock Improving the perceptual quality of 2d animation interpolation.
\newblock In {\em Computer Vision -- ECCV 2022}, Lecture Notes in Computer Science, pages 271--287. Springer, 2022.

\bibitem{chen2024seine}
Xinyuan Chen, Yaohui Wang, Lingjun Zhang, Shaobin Zhuang, Xin Ma, Jiashuo Yu, Yali Wang, Dahua Lin, Yu Qiao, and Ziwei Liu.
\newblock {SEINE}: Short-to-long video diffusion model for generative transition and prediction.
\newblock In {\em International Conference on Learning Representations (ICLR)}. OpenReview.net, 2024.

\bibitem{danier2024ldmvfi}
Duolikun Danier, Fan Zhang, and David Bull.
\newblock {LDMVFI}: Video frame interpolation with latent diffusion models.
\newblock In {\em Proceedings of the AAAI Conference on Artificial Intelligence}, volume~38, pages 1472--1480. AAAI Press, 2024.

\bibitem{guo2024sparsectrl}
Yuwei Guo, Ceyuan Yang, Anyi Rao, Maneesh Agrawala, Dahua Lin, and Bo Dai.
\newblock Sparsectrl: Adding sparse controls to text-to-video diffusion models.
\newblock In {\em Computer Vision -- ECCV 2024}, Lecture Notes in Computer Science, pages 330--348. Springer, 2024.

\bibitem{guo2024animatediff}
Yuwei Guo, Ceyuan Yang, Anyi Rao, Zhengyang Liang, Yaohui Wang, Yu Qiao, Maneesh Agrawala, Dahua Lin, and Bo Dai.
\newblock Animatediff: Animate your personalized text-to-image diffusion models without specific tuning.
\newblock In {\em International Conference on Learning Representations (ICLR)}. OpenReview.net, 2024.

\bibitem{heusel2017fid}
Martin Heusel, Hubert Ramsauer, Thomas Unterthiner, Bernhard Nessler, and Sepp Hochreiter.
\newblock Gans trained by a two time-scale update rule converge to a local nash equilibrium.
\newblock In {\em Advances in Neural Information Processing Systems (NeurIPS)}, volume~30. Curran Associates, Inc., 2017.

\bibitem{ho2020ddpm}
Jonathan Ho, Ajay Jain, and Pieter Abbeel.
\newblock Denoising diffusion probabilistic models.
\newblock In {\em Advances in Neural Information Processing Systems (NeurIPS)}, volume~33, pages 6840--6851. Curran Associates, Inc., 2020.

\bibitem{ho2022videodiffusion}
Jonathan Ho, Tim Salimans, Alexey Gritsenko, William Chan, Mohammad Norouzi, and David~J. Fleet.
\newblock Video diffusion models.
\newblock In {\em Advances in Neural Information Processing Systems (NeurIPS)}, volume~35, pages 8633--8646. Curran Associates, Inc., 2022.

\bibitem{hu2024animateanyone}
Li Hu.
\newblock Animate anyone: Consistent and controllable image-to-video synthesis for character animation.
\newblock In {\em Proceedings of the IEEE/CVF Conference on Computer Vision and Pattern Recognition (CVPR)}, pages 8153--8163, 2024.

\bibitem{huang2024lvcd}
Zhitong Huang, Mohan Zhang, and Jing Liao.
\newblock {LVCD}: Reference-based lineart video colorization with diffusion models.
\newblock {\em ACM Transactions on Graphics}, 43(6), November 2024.

\bibitem{huang2022rife}
Zhewei Huang, Tianyuan Zhang, Wen Heng, Boxin Shi, and Shuchang Zhou.
\newblock Real-time intermediate flow estimation for video frame interpolation.
\newblock In {\em Computer Vision -- ECCV 2022}, Lecture Notes in Computer Science, pages 624--642. Springer, 2022.

\bibitem{jain2024vidim}
Siddhant Jain, Daniel Watson, Eric Tabellion, Aleksander Ho?ynski, Ben Poole, and Janne Kontkanen.
\newblock Video interpolation with diffusion models.
\newblock In {\em Proceedings of the IEEE/CVF Conference on Computer Vision and Pattern Recognition (CVPR)}, pages 7341--7351, June 2024.

\bibitem{jiang2018superslomo}
Huaizu Jiang, Deqing Sun, Varun Jampani, Ming-Hsuan Yang, Erik Learned-Miller, and Jan Kautz.
\newblock Super slomo: High quality estimation of multiple intermediate frames for video interpolation.
\newblock In {\em Proceedings of the IEEE Conference on Computer Vision and Pattern Recognition (CVPR)}, pages 9000--9008, 2018.

\bibitem{jiang2026vidsketch}
Lifan Jiang, Shuang Chen, Boxi Wu, Deng Cai, and Jiahui Zhang.
\newblock Vidsketch: Hand-drawn sketch-driven video generation with diffusion control.
\newblock {\em Neural Networks}, 196:108465, 2026.

\bibitem{khachatryan2023text2videozero}
Levon Khachatryan, Andranik Movsisyan, Vahram Tadevosyan, Roberto Henschel, Zhangyang Wang, Shant Navasardyan, and Humphrey Shi.
\newblock Text2video-zero: Text-to-image diffusion models are zero-shot video generators.
\newblock In {\em Proceedings of the IEEE/CVF International Conference on Computer Vision (ICCV)}, pages 15954--15964, 2023.

\bibitem{kingma2014vae}
Diederik~P. Kingma and Max Welling.
\newblock Auto-encoding variational bayes.
\newblock In {\em International Conference on Learning Representations (ICLR)}, 2014.

\bibitem{kong2022ifrnet}
Lingtong Kong, Boyuan Jiang, Donghao Luo, Wenqing Chu, Xiaoming Huang, Ying Tai, Chengjie Wang, and Jie Yang.
\newblock {IFRNet}: Intermediate feature refine network for efficient frame interpolation.
\newblock In {\em Proceedings of the IEEE/CVF Conference on Computer Vision and Pattern Recognition (CVPR)}, pages 1969--1978, 2022.

\bibitem{lee2020adacof}
Hyeongmin Lee, Taeoh Kim, Tae-Young Chung, Daehyun Pak, Yukyung Ban, and Sangyoun Lee.
\newblock {AdaCoF}: Adaptive collaboration of flows for video frame interpolation.
\newblock In {\em Proceedings of the IEEE/CVF Conference on Computer Vision and Pattern Recognition (CVPR)}, pages 5316--5325, 2020.

\bibitem{li2023amt}
Zhen Li, Zuo-Liang Zhu, Ling-Hao Han, Qibin Hou, Chun-Le Guo, and Ming-Ming Cheng.
\newblock Amt: All-pairs multi-field transforms for efficient frame interpolation.
\newblock In {\em Proceedings of the IEEE/CVF Conference on Computer Vision and Pattern Recognition (CVPR)}, pages 9801--9810, June 2023.

\bibitem{lin2025ctrladapter}
Han Lin, Jaemin Cho, Abhay Zala, and Mohit Bansal.
\newblock Ctrl-adapter: An efficient and versatile framework for adapting diverse controls to any diffusion model.
\newblock In {\em International Conference on Learning Representations (ICLR)}. OpenReview.net, 2025.

\bibitem{liu2025sketchvideo}
Feng-Lin Liu, Hongbo Fu, Xintao Wang, Weicai Ye, Pengfei Wan, Di Zhang, and Lin Gao.
\newblock Sketchvideo: Sketch-based video generation and editing.
\newblock In {\em Proceedings of the Computer Vision and Pattern Recognition Conference (CVPR)}, pages 23379--23390, June 2025.

\bibitem{liu2025manganinja}
Zhiheng Liu, Ka~Leong Cheng, Xi Chen, Jie Xiao, Hao Ouyang, Kai Zhu, Yu Liu, Yujun Shen, Qifeng Chen, and Ping Luo.
\newblock Manganinja: Line art colorization with precise reference following.
\newblock In {\em Proceedings of the Computer Vision and Pattern Recognition Conference (CVPR)}, pages 5666--5677, June 2025.

\bibitem{loftsdottir2022sketchbetween}
Dagmar~Lukka Loftsd{\'o}ttir and Matthew Guzdial.
\newblock Sketchbetween: Video-to-video synthesis for sprite animation via sketches.
\newblock In {\em Proceedings of the 17th International Conference on the Foundations of Digital Games}, pages 32:1--32:7. ACM, 2022.

\bibitem{meng2025anidoc}
Yihao Meng, Hao Ouyang, Hanlin Wang, Qiuyu Wang, Wen Wang, Ka~Leong Cheng, Zhiheng Liu, Yujun Shen, and Huamin Qu.
\newblock Anidoc: Animation creation made easier.
\newblock In {\em Proceedings of the Computer Vision and Pattern Recognition Conference (CVPR)}, pages 18187--18197, June 2025.

\bibitem{niklaus2017sepconv}
Simon Niklaus, Long Mai, and Feng Liu.
\newblock Video frame interpolation via adaptive separable convolution.
\newblock In {\em Proceedings of the IEEE International Conference on Computer Vision (ICCV)}, pages 261--270, 2017.

\bibitem{pan2024sakuga42m}
Zhenglin Pan.
\newblock Sakuga-42m dataset: Scaling up cartoon research.
\newblock {\em arXiv preprint arXiv:2405.07425}, 2024.

\bibitem{pyscenedetect}
{PySceneDetect Authors}.
\newblock {PySceneDetect}.
\newblock \url{https://github.com/Breakthrough/PySceneDetect}, 2023.
\newblock Accessed: October 1, 2023.

\bibitem{radford2021clip}
Alec Radford, Jong~Wook Kim, Chris Hallacy, Aditya Ramesh, Gabriel Goh, Sandhini Agarwal, Girish Sastry, Amanda Askell, Pamela Mishkin, Jack Clark, Gretchen Krueger, and Ilya Sutskever.
\newblock Learning transferable visual models from natural language supervision.
\newblock In {\em Proceedings of the 38th International Conference on Machine Learning (ICML)}, Proceedings of Machine Learning Research, pages 8748--8763. PMLR, 2021.

\bibitem{reda2022film}
Fitsum~A. Reda, Janne Kontkanen, Eric Tabellion, Deqing Sun, Caroline Pantofaru, and Brian Curless.
\newblock {FILM}: Frame interpolation for large motion.
\newblock In {\em Computer Vision -- ECCV 2022}, Lecture Notes in Computer Science, pages 250--266. Springer, 2022.

\bibitem{rombach2022ldm}
Robin Rombach, Andreas Blattmann, Dominik Lorenz, Patrick Esser, and Bj\"orn Ommer.
\newblock High-resolution image synthesis with latent diffusion models.
\newblock In {\em Proceedings of the IEEE/CVF Conference on Computer Vision and Pattern Recognition (CVPR)}, pages 10684--10695, June 2022.

\bibitem{shi2023lineartcolorization}
Min Shi, Jia-Qi Zhang, Shu-Yu Chen, Lin Gao, Yu-Kun Lai, and Fang-Lue Zhang.
\newblock Reference-based deep line art video colorization.
\newblock {\em IEEE Transactions on Visualization and Computer Graphics}, 29(6):2965--2979, 2023.

\bibitem{li2021animeinterp}
Li Siyao, Shiyu Zhao, Weijiang Yu, Wenxiu Sun, Dimitris Metaxas, Chen~Change Loy, and Ziwei Liu.
\newblock Deep animation video interpolation in the wild.
\newblock In {\em Proceedings of the IEEE/CVF Conference on Computer Vision and Pattern Recognition (CVPR)}, pages 6587--6595, June 2021.

\bibitem{song2020ddim}
Jiaming Song, Chenlin Meng, and Stefano Ermon.
\newblock Denoising diffusion implicit models.
\newblock In {\em International Conference on Learning Representations (ICLR)}. OpenReview.net, 2021.

\bibitem{unterthiner2019fvd}
Thomas Unterthiner, Sjoerd van Steenkiste, Karol Kurach, Rapha{\"e}l Marinier, Marcin Michalski, and Sylvain Gelly.
\newblock {FVD}: A new metric for video generation.
\newblock In {\em ICLR Workshop on Deep Generative Models for Highly Structured Data}. OpenReview.net, 2019.

\bibitem{vaswani2017attention}
Ashish Vaswani, Noam Shazeer, Niki Parmar, Jakob Uszkoreit, Llion Jones, Aidan~N. Gomez, {\L}ukasz Kaiser, and Illia Polosukhin.
\newblock Attention is all you need.
\newblock In {\em Advances in Neural Information Processing Systems (NeurIPS)}, volume~30. Curran Associates, Inc., 2017.

\bibitem{voleti2022mcvd}
Vikram Voleti, Alexia Jolicoeur-Martineau, and Chris Pal.
\newblock {MCVD}: Masked conditional video diffusion for prediction, generation, and interpolation.
\newblock In {\em Advances in Neural Information Processing Systems (NeurIPS)}, volume~35. Curran Associates, Inc., 2022.

\bibitem{wang2023videocomposer}
Xiang Wang, Hangjie Yuan, Shiwei Zhang, Dayou Chen, Jiuniu Wang, Yingya Zhang, Yujun Shen, Deli Zhao, and Jingren Zhou.
\newblock Videocomposer: Compositional video synthesis with motion controllability.
\newblock In {\em Advances in Neural Information Processing Systems (NeurIPS)}, volume~36. Curran Associates, Inc., 2023.

\bibitem{wang2025generativeinbetweening}
Xiaojuan Wang, Boyang Zhou, Brian Curless, Ira Kemelmacher-Shlizerman, Aleksander Holynski, and Steven~M. Seitz.
\newblock Generative inbetweening: Adapting image-to-video models for keyframe interpolation.
\newblock In {\em International Conference on Learning Representations (ICLR)}. OpenReview.net, 2025.

\bibitem{wang2004ssim}
Zhou Wang, Alan~C. Bovik, Hamid~R. Sheikh, and Eero~P. Simoncelli.
\newblock Image quality assessment: From error visibility to structural similarity.
\newblock {\em IEEE Transactions on Image Processing}, 13(4):600--612, 2004.

\bibitem{wang2024motionctrl}
Zhouxia Wang, Ziyang Yuan, Xintao Wang, Yaowei Li, Tianshui Chen, Menghan Xia, Ping Luo, and Ying Shan.
\newblock Motionctrl: A unified and flexible motion controller for video generation.
\newblock In {\em ACM SIGGRAPH 2024 Conference Papers}. Association for Computing Machinery, 2024.

\bibitem{wu2023tuneavideo}
Jay~Zhangjie Wu, Yixiao Ge, Xintao Wang, Stan~Weixian Lei, Yuchao Gu, Yufei Shi, Wynne Hsu, Ying Shan, Xiaohu Qie, and Mike~Zheng Shou.
\newblock Tune-a-video: One-shot tuning of image diffusion models for text-to-video generation.
\newblock In {\em Proceedings of the IEEE/CVF International Conference on Computer Vision (ICCV)}, pages 7623--7633, 2023.

\bibitem{xing2024tooncrafter}
Jinbo Xing, Hanyuan Liu, Menghan Xia, Yong Zhang, Xintao Wang, Ying Shan, and Tien-Tsin Wong.
\newblock Tooncrafter: Generative cartoon interpolation.
\newblock {\em ACM Transactions on Graphics}, 43(6), November 2024.

\bibitem{xing2024dynamicrafter}
Jinbo Xing, Menghan Xia, Yong Zhang, Hao Chen, Wangbo Yu, Hanyuan Liu, Gongye Liu, Xintao Wang, Ying Shan, and Tien-Tsin Wong.
\newblock Dynamicrafter: Animating open-domain images with video diffusion priors.
\newblock In {\em Computer Vision -- ECCV 2024}, Lecture Notes in Computer Science, pages 399--417. Springer, 2024.

\bibitem{xu2024magicanimate}
Zhongcong Xu, Jianfeng Zhang, Jun~Hao Liew, Hanshu Yan, Jia-Wei Liu, Chenxu Zhang, Jiashi Feng, and Mike~Zheng Shou.
\newblock Magicanimate: Temporally consistent human image animation using diffusion model.
\newblock In {\em Proceedings of the IEEE/CVF Conference on Computer Vision and Pattern Recognition (CVPR)}, pages 1481--1490, 2024.

\bibitem{zhang2021sketchmeavideo}
Haichao Zhang, Gang Yu, Tao Chen, and Guozhong Luo.
\newblock Sketch me a video.
\newblock {\em arXiv preprint arXiv:2110.04710}, 2021.

\bibitem{zhang2023controlnet}
Lvmin Zhang, Anyi Rao, and Maneesh Agrawala.
\newblock Adding conditional control to text-to-image diffusion models.
\newblock In {\em Proceedings of the IEEE/CVF International Conference on Computer Vision (ICCV)}, pages 3836--3847, October 2023.

\bibitem{zhang2018lpips}
Richard Zhang, Phillip Isola, Alexei~A. Efros, Eli Shechtman, and Oliver Wang.
\newblock The unreasonable effectiveness of deep features as a perceptual metric.
\newblock In {\em Proceedings of the IEEE Conference on Computer Vision and Pattern Recognition (CVPR)}, June 2018.

\bibitem{zhu2025tps}
Tianyi Zhu, Wei Shang, and Dongwei Ren.
\newblock Thin-plate spline-based interpolation for animation line inbetweening.
\newblock In {\em Proceedings of the AAAI Conference on Artificial Intelligence}, volume~39, pages 10995--11003. AAAI Press, 2025.

\end{thebibliography}
}

\end{document}